\documentclass[10pt,twocolumn,letterpaper]{article}

\usepackage[applications]{wacv}      

\definecolor{wacvblue}{rgb}{0.21,0.49,0.74}
\usepackage[pagebackref,breaklinks,colorlinks,allcolors=wacvblue]{hyperref}

\usepackage{graphicx}
\usepackage{amsmath} 
\usepackage{mathtools}
\usepackage{bm}
\usepackage{multirow}
\usepackage{amsthm} 
\usepackage{algorithm}
\usepackage{algorithmic}
\usepackage{xcolor}
\usepackage{pifont}
\usepackage{caption}
\usepackage{soul}
\usepackage{array}

\newtheorem{Proposition_proof}{Proposition}

\newcommand{\cmark}{\ding{51}}%
\newcommand{\xmark}{\ding{55}}%

\usepackage{amsmath,amsfonts,bm}

\def\eqref#1{equation~\ref{#1}}

\def\1{\bm{1}}

\DeclareMathAlphabet{\mathsfit}{\encodingdefault}{\sfdefault}{m}{sl}
\SetMathAlphabet{\mathsfit}{bold}{\encodingdefault}{\sfdefault}{bx}{n}

\DeclareMathOperator*{\argmin}{arg\,min}

\def\wacvPaperID{308} 
\def\confName{WACV}
\def\confYear{2026}

\title{DODA: Adapting Object Detectors to Dynamic Agricultural Environments in Real-Time with Diffusion}

\author{Shuai Xiang, Pieter M. Blok, James Burridge, Haozhou Wang \& Wei Guo \thanks{Corresponding Author} \\
Graduate School of Agricultural and Life Sciences\\
The University of Tokyo\\
1-1-1 Midori-cho, Nishitokyo City\\
\footnotesize{\texttt{\{xiang-shuai, pieterblok, burridge-j, haozhou-wang, guowei\}@g.ecc.u-tokyo.ac.jp}} \\
}

\begin{document}
\maketitle
\begin{abstract}
Object detection has wide applications in agriculture, but domain shifts of diverse environments limit the broader use of the trained models. Existing domain adaptation methods usually require reapplying the entire method for new domains, which is impractical for agricultural applications due to constantly changing environments. In this paper, we propose DODA (\underline{D}iffusion for \underline{O}bject-detection \underline{D}omain Adaptation in \underline{A}griculture), a diffusion-based framework that can adapt the detector to a new domain in just 2 minutes (on one Nvidia 4090). DODA incorporates external domain embeddings and an improved layout-to-image approach, allowing it to generate high-quality detection data for new domains without additional training. We demonstrate DODA's effectiveness on the Global Wheat Head Detection dataset, where fine-tuning detectors on DODA-generated data yields significant improvements across multiple domains. DODA provides a simple yet powerful solution for agricultural domain adaptation, reducing the barriers for growers to use detection in personalized environments.
\end{abstract}    
\section{Introduction}
\label{sec:intro}

Object detection has been widely used in various aspects of agriculture, such as yield estimation \cite{wang2022fast, wang2022field}, disease identification \cite{wu2021application, zhang2020deep}, and decision-making support \cite{bazame2021detection, wang2023drone}. These applications can improve the efficiency and profitability of plant breeders and farmers to improve their efficiency and profits. These models are built for specific farms, crop varieties and management systems, and excel primarily in their specific settings. However, agricultural scenarios are very diverse, the domain shifts caused by factors such as crop varieties, growth stages, cultivation management, and imaging pipeline, making directly applying a given model to new environments unfeasible.

The task of overcoming domain shifts between source and target domains is known as domain adaptation (DA) \cite{wang2018deep}. Some studies \cite{ma2021adaptive, zhang2021easy} have successfully applied existing DA methods to agriculture. However, most of existing DA methods are designed for fixed target domains (e.g., from day to night \cite{kennerley20232pcnet, zhang2024isp}, or from clear to foggy weather \cite{Gao_2023_CVPR, khodabandeh2019robust}) and thus usually require retraining the entire model for other new domains. This limitation poses particular challenges in agriculture, where environmental conditions vary significantly across users. Each time a user applies the model, DA may be necessary. Thus, developing a fast and user-friendly DA method is crucial for practical agricultural applications, enabling models or services to be adapted efficiently to diverse environments without heavy computational demands or technical expertise.

Fine-tuning the model using data from the target domain is a fast adaptation approach. However, the manual effort required for data annotation makes this impractical. Recently, diffusion models \cite{DDPM, rombach2022high} have attracted attention for their ability to rapidly generate high-quality, novel images that are not present in the training set. A growing number of studies explore the potential for diffusion models to address data-related challenges, including visual representation learning \cite{tian2024stablerep, tian2023learning}, classification \cite{azizi2023synthetic, sariyildiz2023fake}, and semantic segmentation \cite{schnell2023generative, tan2023diffss, xie2023mosaicfusion}. For diffusion-based detection data generation, the existing methods can be categorized into three types. 1. Copy-paste Synthesis \cite{ge2022dall, lin2023explore}: foreground and background are synthesized separately and then combined, which often results poor image consistency. 2. Direct Image Generation \cite{zhang2023diffusionengine, feng2024instagen}: images are generated using a text-to-image model, with labels derived from a detector or module. While this approach ensures consistency, it is unsuitable for DA, as the detector often fails in unseen domains. 3. Layout-to-Image (L2I) Generation \cite{chen2023integrating, zheng2023layoutdiffusion, cheng2023layoutdiffuse}: uses semantic layouts to control the spatial arrangement of generated objects. While accurate and effective, it requires image-layout pairs for training, making the generated data mainly useful for enhancing its own training set rather than DA. These raise the question: \textit{How can diffusion model be leveraged to generate high-quality detection data for new domains?}

\begin{figure*}[t]
  \centering
  \includegraphics[width=\linewidth]{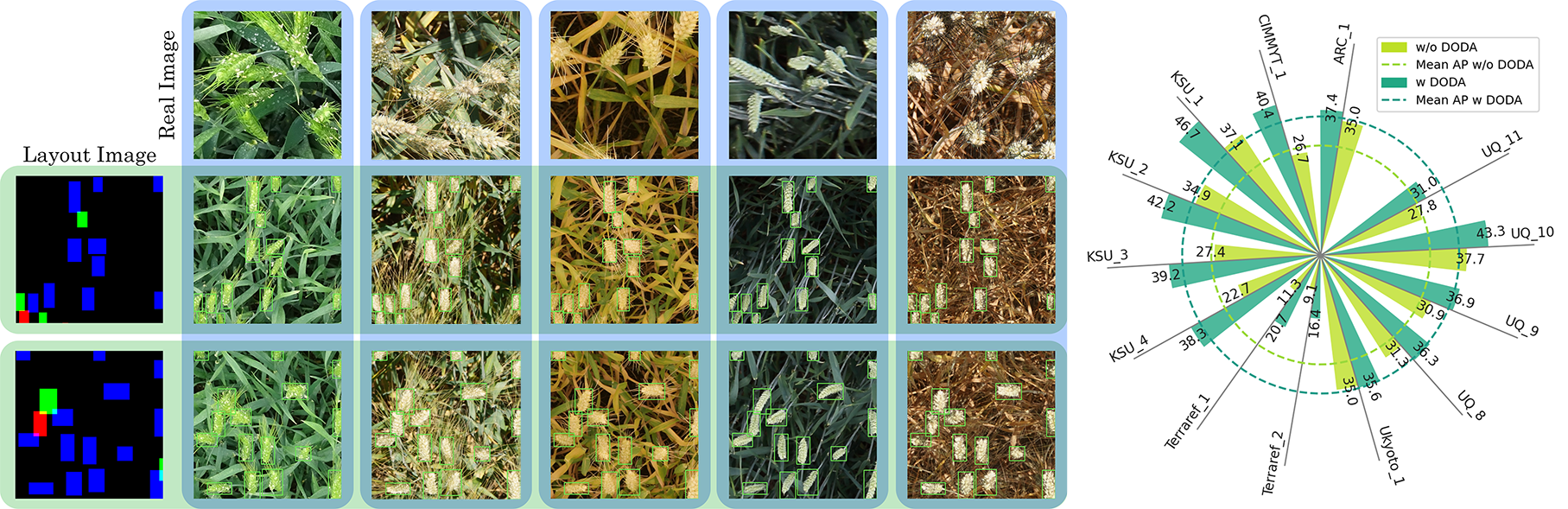}
  \caption{Overview. Left, We propose \textbf{DODA} to generate detection data for diverse agricultural domains, the context of the generated images matches the target domain, and the layout of the generated images aligns with the input layout images. Right, fine-tuning detector on DODA-generated data yields significant improvements across multiple domains.}
  \label{fig:Overview}
\end{figure*}

To address the challenge of diverse and unknown domains in agricultural applications, we propose DODA, a unified framework for generating detection data in real-time across diverse domains. First, we introduce a new L2I method to enhance control over the layout of generated images, ensuring label accuracy. Our L2I method achieves 42.5 mAP on the COCO dataset \cite{lin2014microsoft}, surpassing the previous SOTA L2I method, GeoDiffusion \cite{chen2023integrating}, by +14.8 mAP, and approaching the performance of real images (45.2 mAP). Second, we condition the diffusion model with external domain embeddings. This design decouples the learning of domain-specific features from the diffusion model, enabling it to generate images for target domains without additional training (Fig.~\ref{fig:Overview} left). Extensive adaption experiments on the Global Wheat Head Detection (GWHD) dataset \cite{david2021global}, the largest agricultural detection dataset spanning diverse sub-domains, demonstrate consistent improvements across multiple domains (as shown in Fig.~\ref{fig:Overview} right). On GWHD, DODA achieves a maximum AP increase of 15.6, with the fastest adaptation taking just 2 minutes. These results indicate that DODA can lower the barriers for growers who use detection in their personalized environments.

The main contributions of this paper can be summarized as:
\begin{itemize}
\item We propose a new L2I method that can better contrl the layout and improves label accuracy. Results on the COCO dataset show it generates more accurate layouts, significantly outperforming previous L2I methods.
\item We propose a decoupled conditioning mechanism that incorporates external domain embeddings, enabling fast adaptation to unseen domains without retraining.
\item We propose dividing the training process into pre-training and post-training, and we demonstrate that scaling pre-training dataset with additional unlabeled images effectively improves the quality of generated data.
\item Our method achieves substantial and consistent improvements in AP across multiple domains on the GWHD dataset, with adaptation completed in 2 minutes (on a consumer-grade GPU NVIDIA 4090.). This highlights the DODA's practicality for rapid and cost-effective domain adaptation in real-world agricultural applications.
\end{itemize}

\section{Related work}
\label{sec:rel_work}

\textbf{Global Wheat Head Detection dataset.} The GWHD \cite{david2021global} dataset is one of the largest agricultural detection datasets, specifically focused on close-range wheat head detection. It consists of 47 sub-domains, each with certain differences, such as location, imaging pipeline, collection time, wheat development stages, and wheat varieties. This division allows the development and evaluation of a robust domain adaptation algorithm that performs well under different agricultural environments.

\noindent \textbf{Domain shift and adaptation.} Domain shift (DS) can be divided into image-level shifts (e.g., lighting and color variations that subtly alter feature distributions) and instance-level shifts (e.g., differences in object pose, category, or position) \cite{ma2022i2f}. Such shifts often degrade model performance when applied to new domains. Few-shot domain adaptation (FDA) \cite{Gao_2023_CVPR, Nakamura_2022_ACCV} addresses this by aligning representations between source data and a small amount of labeled target data, while unsupervised domain adaptation (UDA) improves performance without target labels. However, most DA methods are designed for fixed target domains and require the full pipeline to be reapplied for new target domains, which limits their applicability in open-world agricultural settings. To overcome these challenges, we leverage generative models to synthesise data. Specifically, we decouple the learning of domain-specific features from the diffusion model by conditioning on external domain information. This enables the generation of data for new domains without the need for retraining, thus allowing detectors to adapt efficiently to novel environments.

\noindent \textbf{Layout-to-image generation.} The category and position of all objects in an image are referred to as the layout. The layout-to-image (L2I) task aims to synthesize images that follow a given layout. Layouts can be represented in two ways: (1) bounding boxes, which specify the position of an object using the coordinates of its four vertices, and (2) masks, which specify both the shape and position through semantic segmentation. Bounding-box–based methods \cite{chen2023integrating, rombach2022high, zheng2023layoutdiffusion} typically use a text encoder to process the coordinates. However, this approach often struggles to align the layout with image features, resulting in weaker control. In contrast, mask-based methods (Mask-to-Image, M2I) \cite{zhang2023adding} provide more precise layout control. Despite this advantage, masks are difficult to generate algorithmically and lack flexibility, making M2I less suitable for data generation. To address these issues, we propose representing bounding boxes in image form (Fig.~\ref{fig:Overview}, left) and encoding them with an image encoder. This design improves the alignment between layout and image features, enhances layout control, and offers greater flexibility.

\section{Method}
\label{sec:method}

\subsection{Preliminaries}
\label{M:pre}
Song et al. \cite{song2020score} provided a perspective on explaining the diffusion model \cite{DDPM} from the stochastic differential equation (SDE) and score-based generative models \cite{song2019sgm, hyvarinen2005estimation}. The forward diffusion process perturbs the data with random noise, described by the following SDE:
\begin{equation}
  \label{eqn:forward_sde}
  d \mathbf{x} = f(\mathbf{x}, t)dt + g(t)\mathbf{w}
\end{equation}
Where the $f: {\mathbb{R}}^d \to {\mathbb{R}}^d$ is the drift coefficient of $\mathbf{x}_t$, $g: \mathbb{R} \to \mathbb{R}$ is the diffusion coefficient of $\mathbf{x}_t$, and $\mathbf{w}$ is the standard Brownian motion.

The forward diffusion process gradually transforms the data from the original distribution $p(\mathbf{x}_0)$ into a simple noise distribution $p(\mathbf{x}_T)$, over time $T$. By reversing this process, we can sample $\mathbf{x}_0 \sim p(\mathbf{x}_0)$ starting from random noise. According to \cite{anderson1982reverse}, this reverse process is given by a reverse-time SDE:
\begin{equation}
  \label{eqn:backward_sde}
  d\mathbf{x} = [f(\mathbf{x}, t) - g(t)^2  \nabla_{\mathbf{x}_t} \log p(\mathbf{x}_t)] dt + g(t) d \bar{\mathbf{w}}
\end{equation}
Where the $\bar{w}$ is the standard Brownian motion in reverse time. In practice, a neural network (Usually a U-Net) $s(x_t, t; {\boldsymbol{\theta}})$ is used to estimate the score \(\nabla_{\mathbf{x}_t} \log p(\mathbf{x}_t)\) for each time step, thereby approximating the reverse SDE. The optimization objective of the model can be written as:
\begin{equation}
  \label{eqn:opt_sgm}
  \begin{aligned}
  {\boldsymbol{\theta}}^* = \argmin_{\boldsymbol{\theta}} \,
  & \mathbb{E}_{t \sim U(0, T)} \,
  \mathbb{E}_{\mathbf{x}_0 \sim p(\mathbf{x}_0)} \,
  \mathbb{E}_{\mathbf{x}_t \sim p(\mathbf{x}_t|\mathbf{x}_0)} \\
  & [\lambda(t)\| s(\mathbf{x}_t, t; {\boldsymbol{\theta}}) 
  - \nabla_{\mathbf{x}_t}\log p(\mathbf{x}_t|\mathbf{x}_0) \|^2 ]
  \end{aligned}
\end{equation}
Where $\lambda: [0, T] \to {\mathbb{R}}_+$ is a weighting function with respect to time, \(\nabla_{\mathbf{x}_t} \log p(\mathbf{x}_t|\mathbf{x}_0)\) can be obtained through the transition kernel of the forward process. 
Given sufficient data and model capacity, the converged model $s(\mathbf{x}_t, t; {{\boldsymbol{\theta}}^*})$ matches $\nabla_{\mathbf{x}_t} \log p(\mathbf{x}_t)$ for almost all $\mathbf{x}_t$ \cite{song2020score}.

\subsection{Problem Formulation}
\label{M:PF}
In this paper, we aim to improve the detector's recognition of new agricultural scenes in agriculture with limited labeled data. Assume $\mathcal{D}^{(1)} = \{\mathbf{x}^{(1)}, \mathbf{y}_1^{(1)}, \mathbf{y}_2^{(1)} \}$ is an existing object detection dataset, where $\mathbf{x}^{(1)}$ represents all the images in $\mathcal{D}^{(1)}$, $\mathbf{y}_1^{(1)}$ and $\mathbf{y}_2^{(1)}$  represents the domain information and the bounding box annotations of $\mathbf{x}^{(1)}$, respectively. $\mathbf{x}^{(2)}$, $\mathbf{y}_1^{(2)}$ are images from the new scenes and their corresponding domain information. Because $ \mathbf{y}_1^{(1)} \neq \mathbf{y}_1^{(2)} $, the detectors trained on $\mathcal{D}^{(1)}$ may not be able to recognize $\mathbf{x}^{(2)}$. We expect to leverage diffusion to build a synthetic dataset $\hat{\mathcal{D}}^{(2)} = \{\hat{\mathbf{x}}^{(2)}, \mathbf{y}_1^{(2)}, \hat{\mathbf{y}}_2^{(2)} \}$, and improve detectors' recognition of $\mathbf{x}^{(2)}$ by fine-tuning on $\hat{\mathcal{D}}^{(2)}$.

First, the images generated by the diffusion model should align with the context of the target domain. This requires the diffusion model to distinguish between different domains and sample from $p(\mathbf{x}|\mathbf{y}_1^{(2)})$. Given that $\hat{\mathbf{x}}^{(2)}$ is expected to resemble $\mathbf{x}^{(2)}$, the common approach for constructing synthetic datasets, synthesizing $\hat{\mathbf{x}} \sim p(\mathbf{x})$ first, then obtaining labels with an off-the-shelf model \cite{li2023open, zhang2023diffusionengine, kim2024sddgr}, cannot be applied. To generate detection data for the new domains, the diffusion model should be able to sample from $p(\mathbf{x} | \mathbf{y}_1^{(2)}, \hat{\mathbf{y}}_2^{(2)})$.

\begin{figure*}[t]
  \centering
  \includegraphics[width=\linewidth]{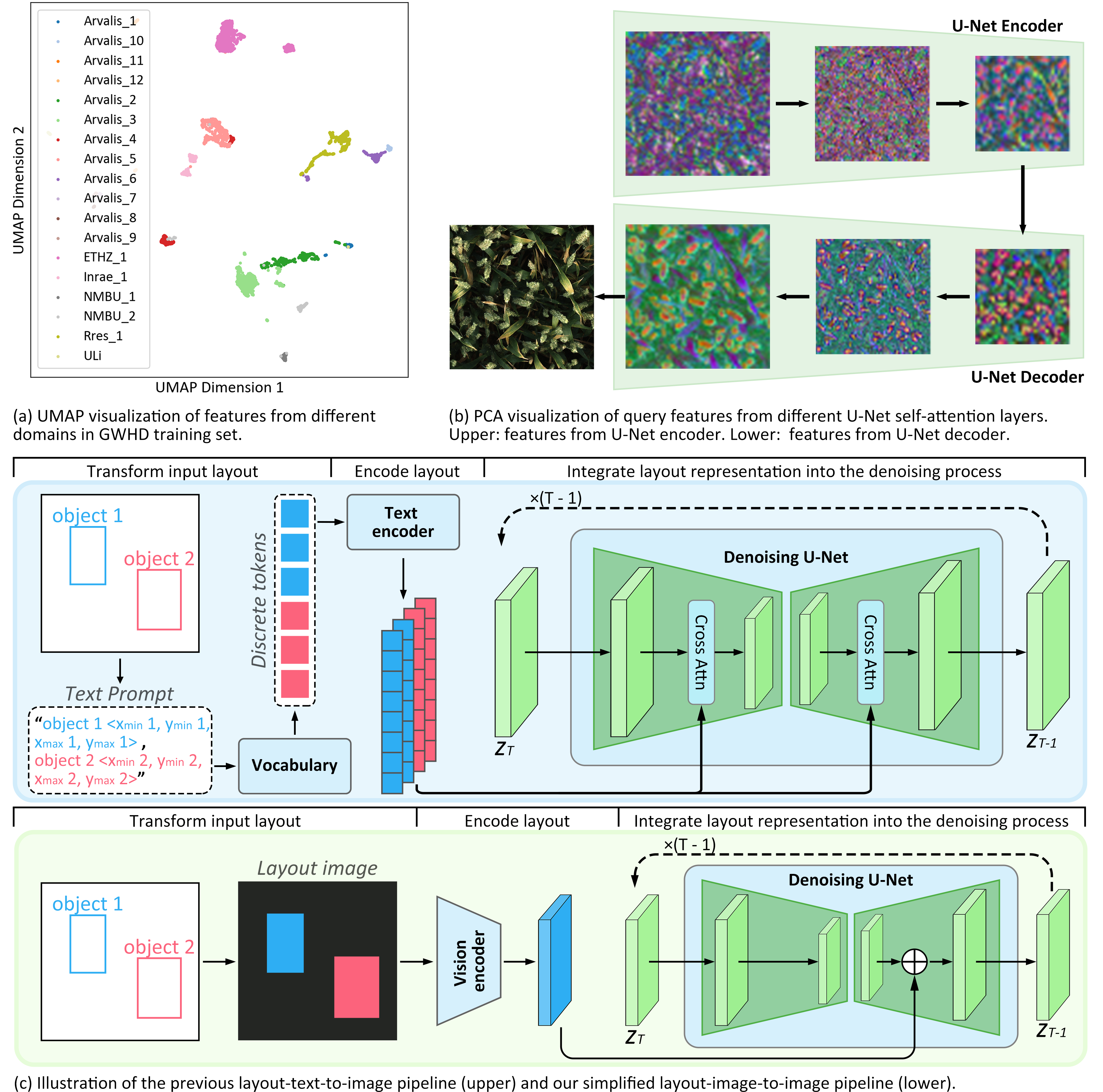}
  \caption{(a) Visualization of the image features from the GWHD training set. The image features are extracted by MAE \cite{he2022masked} and different subdomains are distinguishable by color. (b) Features in shallow layers are relatively noisy, while deeper layers progressively form a clearer layout of the image. (c) Pipelines of existing text-based L2I methods and our LI2I method, by simplifying existing methods, LI2I can better retain spatial information and integrate layout features into U-Net.}
  \label{fig:method}
\end{figure*}

\subsection{DODA}
\begin{figure}[t]
  \centering
  \includegraphics[width=\linewidth]{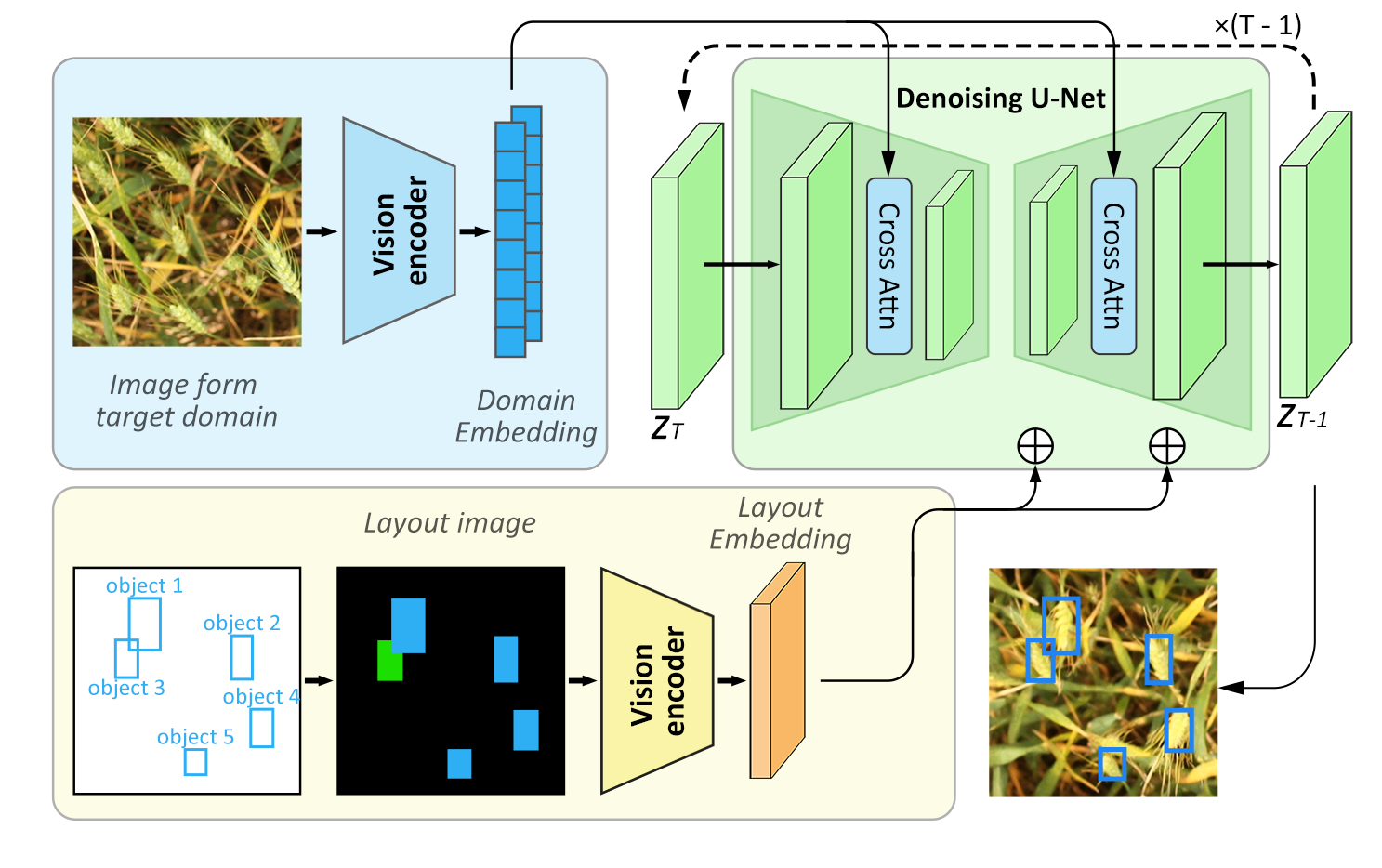}
  \caption{The architecture of DODA: Blue part is the domain encoder, which extracts domain embeddings from the reference image and guide the style of generated image. The orange part is the layout encoder, which encodes the layout image into a feature map and guide the layout.}
  \label{fig:structure}
\end{figure}

The overall architecture of our DODA is shown in Fig.~\ref{fig:structure}. 
\subsubsection{Incorporating Domain Embedding for Domain-Aware Image Generation}
\label{M:DA}
We expect to incorporate external domain embeddings to decouple the learning of domain-specific features from diffusion, thus enabling domain-aware image generation. These domain embeddings should satisfy two key criteria: 1. They must effectively guide the diffusion model to generate images that align with the target domain. This alignment should be at both the image and instance level. 2. They should be easily obtainable for various unseen domains.

\cite{ma2022i2f} demonstrated that combined instance and image level DA produces better results than either method alone. This suggests that the features extracted by the model encompass both image-level and instance-level domain characteristics. Furthermore, \cite{david2021global} used a ResNet \cite{he2016deep} trained on ImageNet to extract features from the GWHD dataset, to suggest that dimensionality reduction can distinguish training and test set features. Similarly, as shown in Fig.~\ref{fig:method}a, our finer-grained test indicates that simple dimensionality reduction can differentiate features from different domains.This suggests that domain-specific features reflect unique domain characteristics, even without prior training on those domains.

Based on the above, we propose using a pre-trained vision encoder to extract features as domain embeddings:
\begin{equation}
  \label{eqn:domain_emb}
  {y}_1^{(2)} = f_{Domain}({x}^{(2)})
\end{equation}
Where the $f_{Domain}: {\mathbb{R}}^{H \times W \times 3} \to {\mathbb{R}}^{d_1}$ is the vision encoder, ${x}^{(2)}$ is the reference image from target domain. The embeddings ${y}_1^{(2)}$ is then integrated into the U-Net via cross-attention:
\begin{equation}
  \label{eqn:domain_att}
  \resizebox{\linewidth}{!}{$
  \begin{aligned}
  \text{Cross-Att}(Q, K, V) 
  &= \text{softmax}\left({QK^\top}{d^{-\tfrac{1}{2}}}\right)V \\ 
  &= \text{softmax}\left({Q(W^K {y}_1^{(2)})^\top}{d^{-\tfrac{1}{2}}}\right)W^V {y}_1^{(2)}
  \end{aligned}
  $}
\end{equation}
Where the $W^K \in {\mathbb{R}}^{d_1 \times d_2}$ and $W^V \in {\mathbb{R}}^{d_1 \times d_2}$ are projection matrices that convert domain embeddings into Key and Value. 
By default, we use a ViT-B \cite{dosovitskiy2020image} pre-trained with CLIP \cite{radford2021learning} as the domain encoder.

\noindent \textbf{Decoupling layout information from domain embedding.} The features extracted by the vision encoder contain both domain-specific features and layout information. The layout information can interfere with controlling the layout of the generated images. To decouple layout information from domain embedding, we employ a simple yet effective method termed asymmetric augmentation. In asymmetric augmentation, both the domain reference image and the denoising target image are obtained by transforming the original image during training, but the augmentation of the reference image is relatively weaker. More details can be found in Appendix \ref{app:imple}. 

\subsubsection{Encoding Layout Images with Vision Model for Simpler and Better Alignment}
\label{sec:layout_encoder}
Generating images from layouts is a key focus in image generation. As shown in Fig.~\ref{fig:method}c upper, existing methods \cite{zheng2023layoutdiffusion, chen2023integrating, rombach2022high, li2023gligen} first represent the layout as text, and encode it with a language model, then fuse the layout embedding into the diffusion model through cross-attention. We refer to these methods collectively as LT2I (layout-text-to-image). From our perspective, LT2I disrupts the layout's spatial relationships in multiple ways: 1. Since, LT2I represents the layout as text, which is embedded as discrete tokens before being fed into the transformer, The transformer is forced to learn to reconstruct the spatial relationships from the discretized tokens. 2. Before cross-attention, features from the diffusion model are flattened, which degrades spatial information and makes alignment more difficult.

We propose to address these problems by simplifying the process. As shown in the lower part of Fig.~\ref{fig:method}c, inspired by \cite{zhang2023adding, mou2024t}, we represent the layout as image, which naturally has accurate spatial information, eliminating the need of learning. The layout encoder then interprets the hierarchical and positional relationships between bounding boxes, and converts rectangles into object shapes. Finally, we fuse the features through addition to maximally retain spatial information. Corresponding to LT2I, we refer to this method as LI2I (layout-image-to-image).

\noindent \textbf{Design of Layout Encoder.} Our layout encoder has a simple structure, as it does not need to convert the discrete input to spatial relationships. The layout encoder consists of a stack of time-dependent residual layers and downsampling layers. The output of each residual layer is $f_{res}(\mathbf{a}, t)+\mathbf{a}$, here $\mathbf{a}$ is the output of last layer, and $t$ is the timestep.

\noindent \textbf{Layers to Merge Layout Embeddings.} We observed that shallow U-Net layers produce noisy, localized features. As the layers deepen, the features become increasingly abstract and holistic, gradually forming the overall layout of the image (shown in Fig.~\ref{fig:method}b). Therefore, we merge the layout embeddings with the deeper layers to better convey the layout information.

\noindent \textbf{Channel Coding for Overlapped Instances.} Bounding boxes will overlap with each other, so to help the layout encoder distinguish instances, we assign different colors to overlapped instances. Specifically, we represent the overlap relationships of the bounding boxes as an adjacency matrix, thereby calculating the overlapping relationships and assigning colors (Algorithm in  Appendix.~\ref{alg:greedycoloring}).

\noindent The Fig.~\ref{fig:method} illustrates the overview of our DODA architecture.

\subsubsection{Unified Optimization Objective for Multi-Conditional Diffusion}
According to Sec.~\ref{M:PF}, our goal is to sample from $p(\mathbf{x}_t | \mathbf{y}_1, \mathbf{y}_2)$, the training objective can be written as:
\begin{equation}
  \label{eqn:opt_sgm_2condition1}
  \resizebox{\linewidth}{!}{$
  \begin{aligned}
  {\boldsymbol{\theta}}^* = \argmin_{\boldsymbol{\theta}} \,
  & \mathbb{E}_{t \sim U(0, T)} \,
  \mathbb{E}_{\mathbf{x}_t,\mathbf{y}_1,\mathbf{y}_2 \sim p(\mathbf{x}_t,\mathbf{y}_1,\mathbf{y}_2)} \\ 
  & [\lambda(t)\| s(\mathbf{x}_t,\mathbf{y}_1,\mathbf{y}_2,t;{\boldsymbol{\theta}}) 
  - \nabla_{\mathbf{x}_t}\log p(\mathbf{x}_t|\mathbf{y}_1,\mathbf{y}_2) \|^2 ]
  \end{aligned}
  $}
\end{equation}

By Proposition~\ref{prop:2condition}, this objective is equivalent to:
\begin{equation}
  \label{eqn:opt_cgm_2condition1}
  \resizebox{\linewidth}{!}{$
    \begin{aligned}
    {\boldsymbol{\theta}}^* = \argmin_{\boldsymbol{\theta}} \,
    & \mathbb{E}_{t \sim U(0, T)} \,
    \mathbb{E}_{\mathbf{x}_0,\mathbf{y}_1,\mathbf{y}_2 \sim p(\mathbf{x}_0,\mathbf{y}_1,\mathbf{y}_2)} \,
    \mathbb{E}_{\mathbf{x}_t \sim p(\mathbf{x}_t|\mathbf{x}_0)} \\ 
    & [\lambda(t)\| s(\mathbf{x}_t,\mathbf{y}_1,\mathbf{y}_2,t;{\boldsymbol{\theta}}) 
    - \nabla_{\mathbf{x}_t}\log p(\mathbf{x}_t|\mathbf{x}_0) \|^2 ]
    \end{aligned}
  $}
\end{equation}
Eq.~\ref{eqn:opt_cgm_2condition1} can be optimized directly and serves as our actual training objective. The complete derivation of training objective is provided in Appendix \ref{app:derivation}.

\noindent \textbf{Two-stage Training.} In practice, the box annotations $\mathbf{y}_2^{(1)}$ in $\mathcal{D}^{(1)}$ are limited, while $\mathbf{x}$ and $\mathbf{y}_1$ are easier to obtain. Therefore, we suggest building a larger dataset without box annotations $\mathcal{D}^{(3)} = \{\mathbf{x}^{(3)}, \mathbf{y}_1^{(3)} \}$, where $\mathbf{x}^{(1)} \subset \mathbf{x}^{(3)}$, $\mathbf{y}_1^{(3)}=f_d(\mathbf{x}^{(3)})$, and perform pre-training on $\mathcal{D}^{(1)}$ to achieve a better estimation of $p(\mathbf{x} | \mathbf{y}_1)$, the training objective can be written as:
\begin{equation}
  \label{eqn:opt_cgm_1condition}
  \begin{aligned}
  {\boldsymbol{\theta}}^* = \argmin_{\boldsymbol{\theta}}
   & \mathbb{E}_{t \sim U(0, T)} 
   \mathbb{E}_{\mathbf{x}_0,\mathbf{y}_1 \sim p(\mathbf{x}_0,\mathbf{y}_1)} 
   \mathbb{E}_{\mathbf{x}_t \sim p(\mathbf{x}_t|\mathbf{x}_0)} \\
   & [\lambda(t)\| s(\mathbf{x}_t,\mathbf{y}_1,t;{\boldsymbol{\theta}}) - \nabla_{\mathbf{x}_t}\log p(\mathbf{x}_t|\mathbf{x}_0) \|^2 ]
  \end{aligned}
\end{equation}
After pre-training, post-training can be conducted using object detection dataset $\mathcal{D}^{(3)}$ with the objective of Eq.~\ref{eqn:opt_cgm_2condition1}. The effectiveness of this two-stage training process is shown in Sec.~\ref{subsec:AB}.
\section{Experiment}
\label{sec:experiment}
By default, DODA employs latent diffusion (LDM) \cite{rombach2022high} as the base diffusion model. The pre-training of DODA is performed on the GWHD 2+ dataset, followed by post-training on the GWHD training set. The implementation details can be found in Appendix \ref{app:imple}. We use Fréchet Inception Distance (FID), Inception Score (IS), COCO metrics, YOLO Score and Feature Similarity (FS) as evaluation metrics. Their specific definitions can be found in Appendix \ref{app:eval_metric}.

\subsection{Main Results}
\subsubsection{Synthetic Data for Agricultural Object Detection Domain Adaptation}
\label{subsec:DA}

We initialize a YOLOX-L \cite{ge2021yolox} model on the GWHD training set, serving as the baseline and base model.

Samples of data generated by DODA for different agricultural domains can be seen in the Fig.~\ref{fig:Overview} left. As shown in Table~\ref{t:domains_adaptation}, the data of 13 domains were collected from different devices, different regions and different stages of wheat head development. After fine-tuning the detector with domain-specific data synthesized by DODA, recognition across these domains improved, with an average improvement of 7.5. 

In addition to the GWHD, we test our method on wheat images collected by UAV, which have significantly different spatial resolutions. As shown in the Table~\ref{t:domains_adaptation}, our method can effectively help the detector adapt to UAV image data. Furthermore, we explore cross-crop adaption. As highlighted in the last row of Table~\ref{t:domains_adaptation}, our method successfully adapts the detector trained on wheat to sorghum \cite{ghosal2019weakly}. These results suggest our method effectively helps the detector adapt to new scenes of agricultural field, bridging the gap between limited manual annotations and complex, ever-changing agricultural environments. Appendix \ref{app:qualitative_da} illustrates more generated data and example of adaption results.

\begin{table*}
  \caption{Domain-specific performance on GWHD test set after fine-tuning with DODA-generated data. Improvements over baseline are marked in \textcolor{ForestGreen}{Green} in parentheses. Consistent improvements across various domains demonstrate that DODA is effective in adapting detectors to new agricultural domains.
  }
  \label{t:domains_adaptation}
  \centering
  \setlength{\tabcolsep}{3.5pt}
  \resizebox{\textwidth}{!}{
    \begin{tabular}{lccccccccc}
    \toprule
    Domain & AP & AP$_{50}$ & AP$_{75}$ & AP$^{s}$ & AP$^{m}$ & AP$^{l}$ & Development stage & Platform & Country\\
    \midrule
    \multicolumn{6}{l}{Global Wheat Head Detection} \\ 
    ARC\_1 & 35.0 & 29.7 & 8.1 & 35.0 & 39.9 & 73.1 & Filling & Handheld & Sudan \\
    \quad+ Ours & 37.4 (\textcolor{ForestGreen}{+2.4}) & 78.3 (\textcolor{ForestGreen}{+5.2}) & 32.3 (\textcolor{ForestGreen}{+2.6}) & 8.3 (\textcolor{ForestGreen}{+0.2}) & 36.8 (\textcolor{ForestGreen}{+1.8}) & 43.0 (\textcolor{ForestGreen}{+3.1}) &&&\\[1pt]
    CIMMYT\_1 & 26.7 & 65.7 & 16.9 & 5.3 & 24.6 & 45.9 & Postﬂowering & Cart & Mexico \\
    \quad+ Ours & 40.4 (\textcolor{ForestGreen}{+13.7}) & 80.7 (\textcolor{ForestGreen}{+15.0}) & 36.1 (\textcolor{ForestGreen}{+19.2}) & 16.2 (\textcolor{ForestGreen}{+10.9}) & 39.2 (\textcolor{ForestGreen}{+14.6}) & 55.5 (\textcolor{ForestGreen}{+9.6}) &&&\\[1pt]
    KSU\_1 & 37.1 & 72.0 & 34.9 & 9.5 & 40.1 & 51.5 & Postﬂowering & Tractor & US \\
    \quad+ Ours & 46.7 (\textcolor{ForestGreen}{+9.6}) & 84.7 (\textcolor{ForestGreen}{+12.7}) & 47.2 (\textcolor{ForestGreen}{+12.3}) & 22.0 (\textcolor{ForestGreen}{+12.5}) & 49.1 (\textcolor{ForestGreen}{+9.0}) & 53.4 (\textcolor{ForestGreen}{+1.9}) &&&\\[1pt]
    KSU\_2 & 34.9 & 74.0 & 29.6 & 6.9 & 38.3 & 58.7 & Postﬂowering & Tractor & US \\
    \quad+ Ours & 42.2 (\textcolor{ForestGreen}{+7.3}) & 86.5 (\textcolor{ForestGreen}{+12.5}) & 33.9 (\textcolor{ForestGreen}{+4.3}) & 16.2 (\textcolor{ForestGreen}{+9.3}) & 44.9 (\textcolor{ForestGreen}{+6.6}) & 61.5 (\textcolor{ForestGreen}{+2.8}) &&&\\[1pt]
    KSU\_3 & 27.4 & 67.8 & 16.2 & 6.0 & 26.5 & 42.6 & Filling & Tractor & US \\
    \quad+ Ours & 39.2 (\textcolor{ForestGreen}{+11.8}) & 81.3 (\textcolor{ForestGreen}{+13.5}) & 31.8 (\textcolor{ForestGreen}{+15.6}) & 16.9 (\textcolor{ForestGreen}{+10.9}) & 39.0 (\textcolor{ForestGreen}{+12.5}) & 50.0 (\textcolor{ForestGreen}{+7.4}) &&&\\[1pt]
    KSU\_4 & 22.7 & 56.3 & 14.6 & 1.2 & 22.6 & 40.6 & Ripening & Tractor & US \\
    \quad+ Ours & 38.3 (\textcolor{ForestGreen}{+15.6}) & 75.1 (\textcolor{ForestGreen}{+18.8}) & 34.1 (\textcolor{ForestGreen}{+19.5}) & 10.0 (\textcolor{ForestGreen}{+8.8}) & 39.3 (\textcolor{ForestGreen}{+16.7}) & 49.4 (\textcolor{ForestGreen}{+8.8}) &&&\\[1pt]
    Terraref\_1 & 11.3 & 33.1 & 5.1 & 1.0 & 13.1 & 43.6 & Ripening & Gantry & US \\
    \quad+ Ours & 20.7 (\textcolor{ForestGreen}{+9.4}) & 54.6 (\textcolor{ForestGreen}{+21.5}) & 10.6 (\textcolor{ForestGreen}{+5.5}) & 4.1 (\textcolor{ForestGreen}{+3.1}) & 24.0 (\textcolor{ForestGreen}{+10.9}) & 43.2 (\textcolor{ForestGreen}{-0.4}) &&&\\[1pt]
    Terraref\_2 & 9.1 & 23.7 & 5.2 & 0.6 & 12.3 & 31.7 & Filling & Gantry & US \\
    \quad+ Ours & 16.4 (\textcolor{ForestGreen}{+7.3}) & 41.5 (\textcolor{ForestGreen}{+17.8}) & 10.2 (\textcolor{ForestGreen}{+5.0}) & 2.7 (\textcolor{ForestGreen}{+2.1}) & 20.7 (\textcolor{ForestGreen}{+8.4}) & 47.2 (\textcolor{ForestGreen}{+15.5}) &&&\\[1pt]
    Ukyoto\_1 & 35.0 & 68.4 & 31.8 & 4.9 & 38.7 & 56.8 & Postﬂowering & Handheld & Japan \\
    \quad+ Ours & 35.6 (\textcolor{ForestGreen}{+0.6}) & 70.1 (\textcolor{ForestGreen}{+1.7}) & 32.6 (\textcolor{ForestGreen}{+0.8}) & 5.5 (\textcolor{ForestGreen}{+0.6}) & 39.0 (\textcolor{ForestGreen}{+0.3}) & 57.3 (\textcolor{ForestGreen}{+0.5}) &&&\\[1pt]
    UQ\_8 & 31.3 & 66.3 & 24.8 & 12.8 & 39.1 & 53.3 & Ripening & Handheld & Australia \\
    \quad+ Ours & 36.3 (\textcolor{ForestGreen}{+5.0}) & 70.3 (\textcolor{ForestGreen}{+4.0}) & 33.6 (\textcolor{ForestGreen}{+8.8}) & 16.6 (\textcolor{ForestGreen}{+3.8}) & 44.4 (\textcolor{ForestGreen}{+5.3}) & 58.3 (\textcolor{ForestGreen}{+5.0}) &&&\\[1pt]
    UQ\_9 & 30.9 & 66.8 & 25.6 & 8.4 & 35.0 & 54.4 & Filling-ripening & Handheld & Australia \\
    \quad+ Ours & 36.9 (\textcolor{ForestGreen}{+6.0}) & 72.3 (\textcolor{ForestGreen}{+5.5}) & 34.8 (\textcolor{ForestGreen}{+9.2}) & 14.6 (\textcolor{ForestGreen}{+6.2}) & 41.0 (\textcolor{ForestGreen}{+6.0}) & 58.4 (\textcolor{ForestGreen}{+4.0}) &&&\\[1pt]
    UQ\_10 & 37.7 & 78.5 & 31.2 & 20.7 & 43.7 & 53.8 & Filling-ripening & Handheld & Australia \\
    \quad+ Ours & 43.3 (\textcolor{ForestGreen}{+5.6}) & 81.5 (\textcolor{ForestGreen}{+3.0}) & 41.1 (\textcolor{ForestGreen}{+9.9}) & 26.5 (\textcolor{ForestGreen}{+5.8}) & 49.0 (\textcolor{ForestGreen}{+5.3}) & 56.6 (\textcolor{ForestGreen}{+2.8}) &&&\\[1pt]
    UQ\_11 & 27.8 & 69.5 & 16.4 & 17.0 & 34.0 & 42.0 & Postﬂowering & Handheld & Australia \\
    \quad+ Ours & 31.0 (\textcolor{ForestGreen}{+3.2}) & 71.6 (\textcolor{ForestGreen}{+2.1}) & 21.6 (\textcolor{ForestGreen}{+5.2}) & 21.2 (\textcolor{ForestGreen}{+4.2}) & 36.2 (\textcolor{ForestGreen}{+2.2}) & 45.7 (\textcolor{ForestGreen}{+3.7}) &&&\\[1pt]
    \midrule
    UAV\_1 & 13.3 & 35.5 & 6.6 & 9.0 & 18.0 & - & Ripening & UAV & - \\
    \quad+ Ours & 28.2 (\textcolor{ForestGreen}{+14.9}) & 54.9 (\textcolor{ForestGreen}{+19.4}) & 25.9 (\textcolor{ForestGreen}{+19.3}) & 20.0 (\textcolor{ForestGreen}{+11.0}) & 37.5 (\textcolor{ForestGreen}{+19.5}) & - &&&\\[1pt]
    Sorghum & 17.3 & 40.0 & 12.6 & 17.7 & 21.5 & - & Ripening & UAV & Australia \\
    \quad+ Ours & 29.4 (\textcolor{ForestGreen}{+12.1}) & 70.5 (\textcolor{ForestGreen}{+30.5}) & 17.9 (\textcolor{ForestGreen}{+5.3}) & 30.0 (\textcolor{ForestGreen}{+12.3}) & 31.5 (\textcolor{ForestGreen}{+10.0}) & - &&&\\[1pt]
    \bottomrule
  \end{tabular}
  }
\end{table*}

\subsubsection{Comparisons with Previous Domain Adaption Methods}
\label{subsec:compaer_DA}
In this section, we compare our method with existing domain adaptation approaches for adapting detectors to the \textbf{Terraref} domain. The \textbf{Terraref} a combination of \textbf{Terraref1} and \textbf{Terraref2}, which are the most challenging domains in the GWHD dataset. As shown in Table~\ref{t:compaer_DA}, our method achieves the best trade-off between accuracy and speed, requiring only 2 minutes for adaptation.

\begin{table}
  \caption{Adaptation results on the \textbf{Terraref} domain. The adaptation time is measured on a single consumer-grade GPU NVIDIA 4090. Source free: source data is not required for adaptation.}
  \label{t:compaer_DA}
  \centering
  \resizebox{\linewidth}{!}{
  \setlength{\extrarowheight}{-1pt}
  \begin{tabular}{lccc}
    \toprule
    Method & Source free & AP\(_{50}\)\(\uparrow\) & Adaptation time\(\downarrow\) \\
    \midrule
    Faster-RCNN \\
    \multicolumn{1}{l|}{Source Only} & - & 18.2 & - \\
    \multicolumn{1}{l|}{SWDA \cite{saito2019strong}} & \xmark & 7.0 & 4.5 hours \\
    \multicolumn{1}{l|}{AASFOD \cite{chu2023adversarial}} & \xmark & 13.7 & 21 hours\\
    \multicolumn{1}{l|}{HTCN \cite{chen2020harmonizing}} & \xmark &  21.7 & 2 hours \\
    \multicolumn{1}{l|}{IRG \cite{vs2023instance}} & \cmark &  22.1 & 1.2 hours \\
    \multicolumn{1}{l|}{MemCLR \cite{vs2023towards}} & \cmark &  22.7 & 3 min \\
    \multicolumn{1}{l|}{DODA (ours)} & \cmark &  \textbf{36.7} & \textbf{2 min} \\
    \midrule
    YOLOv5-X \\
    \multicolumn{1}{l|}{Source Only} & - & 38.6 & - \\
    \multicolumn{1}{l|}{AsyFOD \cite{Gao_2023_CVPR}} & \xmark &  47.0 & 2.5 hours \\
    \multicolumn{1}{l|}{AcroFOD \cite{gao2022acrofod}} & \xmark &  50.5 & 2.5 hours \\
    \multicolumn{1}{l|}{DODA (ours)} & \cmark &  \textbf{53.6} & \textbf{2 min} \\
    \bottomrule
  \end{tabular}
  }
\end{table}

\subsubsection{Comparisons with Previous Layout-to-image Methods}
\label{subsec:L2I_coco}

\begin{table}
  \caption{L2I results on COCO-val2017. Our method achieves significant improvements in controllability compared to previous works. * indicates filtering of objects whose area ratio is less than 0.02 and images with more than 8 objects.}
  \label{t:coco-L2I}
  \centering
  
  \setlength{\tabcolsep}{2pt}
  \resizebox{\linewidth}{!}{
    \begin{tabular}{lcccccccc}
    \toprule
    \multirow{2}{*}{Method} & \multicolumn{6}{c}{YOLO Score\(\uparrow\)} & \multirow{2}{*}{FID\(\downarrow\)} & \multirow{2}{*}{IS\(\uparrow\)} \\
    \addlinespace[-0.8ex]
    \cmidrule(lr){2-7}
    & mAP & AP\(_{50}\) & AP\(_{75}\) & AP\(^{s}\) & AP\(^{m}\) & AP\(^{l}\) \\
    \addlinespace[-0.8ex]
    \midrule
    \(256 \times 256\) \\  
    \multicolumn{1}{l|}{Real image*} & 55.5  & 70.7  & 60.8  & - & 51.2  & 69.0  & - & - \\
    \multicolumn{1}{l|}{PLGAN* \cite{wang2022interactive}} & 21.4 & 35.2  & 22.9   & -   & 16.8   & 27.3   & 35.9  & 17.7\(\pm\)0.9 \\  
    \multicolumn{1}{l|}{LostGANv2* \cite{sun2021learning}} & 26.6 & 41.6  & 28.3   & -   & 21.9   & 34.3   & 37.0  & 17.0\(\pm\)0.9 \\  
    \multicolumn{1}{l|}{LAMA* \cite{li2021image}} & 38.3 & 53.9  & 42.3   & -   & 34.8   & 45.0   & 37.5  & 18.4\(\pm\)1.0 \\  
    \multicolumn{1}{l|}{LayoutDiffusion* \cite{zheng2023layoutdiffusion}} & 30.6 & 56.6  & 29.5   & -   & 20.0   & 43.4   & \textbf{23.6}  & 24.3\(\pm\)1.2 \\
    \multicolumn{1}{l|}{\textbf{LI2I (ours)}*} & \textbf{54.3} & \textbf{72.1}  & \textbf{59.1}  & -  & \textbf{48.7}  & \textbf{59.9}    & 31.5 & \textbf{24.6\(\pm\)1.2} \\
    \cmidrule(lr){2-9}
    \multicolumn{1}{l|}{Real image} & 35.5  & 51.2  & 37.5  & 15.3 & 48.3  & 62.2  & - & 29.0\(\pm\)1.3 \\
    \multicolumn{1}{l|}{LayoutDiffusion \cite{zheng2023layoutdiffusion}} & 6.0 & 14.9  & 3.8   & 0.2   & 4.7   & 19.8   & \textbf{20.5}  & 21.8\(\pm\)1.1 \\
    \multicolumn{1}{l|}{GeoDiffusion \cite{chen2023integrating}} & 27.3  & 38.5   & 29.3  & 2.8   & 40.3  & \textbf{63.2}  & 34.3  & 24.7\(\pm\)1.1 \\
    \multicolumn{1}{l|}{ControlNet M2I \cite{zhang2023adding}} & 29.4  & 39.4   & 31.1  & 11.2   & 39.3  & 52.4  & 49.1  & 18.4\(\pm\)0.9 \\
    \multicolumn{1}{l|}{\textbf{LI2I (ours)}} & \textbf{31.8} & \textbf{45.4}  & \textbf{33.0}  & \textbf{12.3}  & \textbf{41.5}  & 57.2    & 29.9  & \textbf{28.5\(\pm\)0.8} \\
    \midrule
    \(512 \times 512\) \\
    \multicolumn{1}{l|}{Real image} & 45.2  & 63.3  & 48.5  & 17.9 & 45.1  & 61.5  & - & 31.5\(\pm\)1.2 \\
    \multicolumn{1}{l|}{Layout diffuse* \cite{cheng2023layoutdiffuse}} & 4.2  & 11.3  & 2.3  & -     & 0.1  & 4.6  & 33.5 & \textbf{29.6\(\pm\)1.1} \\
    \multicolumn{1}{l|}{GeoDiffusion \cite{chen2023integrating}} & 27.7  & 40.7  & 29.6  & 0     & 13.0  & 57.8  & 28.8  & 26.4\(\pm\)2.4 \\
    \multicolumn{1}{l|}{ControlNet M2I \cite{zhang2023adding}} & 41.0  & 51.7   & 43.6  & 16.0   & 39.0  & 56.5  & 33.9  & 21.3\(\pm\)0.8 \\
    \multicolumn{1}{l|}{\textbf{LI2I (ours)}} & \textbf{42.5}  & \textbf{56.1}  & \textbf{44.9}  & \textbf{16.1}  & \textbf{40.9}  & \textbf{59.1}  & \textbf{24.9}  & 29.4\(\pm\)1.1 \\
    \bottomrule
  \end{tabular}
  }
\end{table}

We evaluate our LI2I architecture on the COCO dataset following prior L2I studies. As shown in Table~\ref{t:coco-L2I}, LI2I achieves clear improvements in controllability (mAP) while maintaining competitive image quality (FID) and diversity (IS). 

\noindent \textbf{Comparison with text-based approaches} (LayoutDiffusion, Layout diffuse, GeoDiffusion). Unlike text-based methods that encode layouts as sequences, LI2I directly represents the layouts as images and processes them with an image encoder. This design avoids the spatial information loss introduced by text discretization. Consequently, LI2I achieves substantially higher mAP and shows particular strength in handling small objects (AP$^s$).

\noindent \textbf{Comparison with mask-based approaches} (ControlNet M2I). ControlNet leverages semantic masks for strong layout control but introduces rigid constraints that often degrade image quality and diversity. In addition, generating accurate masks automatically is non-trivial. By contrast, LI2I offers a more flexible and practical solution for data generation.

\noindent \textbf{Comparison with real image}. The YOLO Score achieved on our generated images closely approximate those obtained from real images. This close alignment highlights the accuracy of our synthetic labels, and indicates significant progress in narrowing the gap between synthetic and real data. 

The quantitative comparison of above methods can be found in Appendix \ref{app:qualitative_comparisons_coco}.

\subsection{Ablation study}
\label{subsec:AB}

In this section, we perform ablation studies to evaluate the impact of each design and setting of the proposed method, additional results are in Appendix \ref{app:more_AB}. 

\noindent \textbf{Scaling dataset and GWHD 2+.} The separation of domain features by the domain encoder enables us to pre-train the diffusion model using a larger set of images, without requiring labels. In Table~\ref{t:pre-training-size}, we explore the effect of the proposed two-stage training. We consider two scenarios: target domain images are accessible (\textbf{w/ target images}) or inaccessible (\textbf{w/o target images}) during pre-training.

As shown in Table~\ref{t:pre-training-size}, when no additional unlabeled data is used, training is one-stage, leading to the worst results. As the size of the pre-training dataset increases, the AP$_{50}$ steadily improves, demonstrating the effectiveness of the two-stage training process. Notably, pre-training diffusion models with unlabeled images from the target domain significantly enhances data quality. To support pre-training, we collect GWHD 2+, an extension of GWHD with 65k additional unlabeled wheat images from 12 domains. As shown in last row of Table~\ref{t:pre-training-size}, GWHD 2+ further improves the data quality and narrows the gap between w/ and w/o access to target images. The performance gains from pre-training with additional unlabeled images, particularly those from the target domain, suggest that our GWHD 2+ dataset is still insufficiently large. A priority for the future is to collect more images to enhance DODA's ability.

\noindent \textbf{Adapting different detectors.} To verify the generalizability of the generated data, we evaluate several detectors with various architectures and sizes. As shown in Table~\ref{t:detectors}, after fine-tuning with domain-specific data generated by DODA, these models show consistent improvement in recognizing \textbf{Terraref}.

\noindent \textbf{Domain encoder.} Table~\ref{t:vision-encoders} shows the effect of the Domain encoder. Without domain encoder, diffusion randomly samples from the training set, leading to lower FS. Independently of the architecture and training data, various pre-trained vision models can guide diffusion to generate images with specific features. Using ViT as the backbone, CLIP performs the best, while MAE \cite{he2022masked} worse than ResNet \cite{he2016deep}. Compared to contrastive learning, MAE focuses more on high-frequency texture features \cite{park2023self, vanyan2023analyzing}, which we hypothesize affects the quality of the domain embedding. To explore this, we test features from different MAE layers, as shallow layers are generally associated with high-frequency, low-level features. As shown in Table~\ref{t:mae-layers}, using shallower MAE features further impacts image quality.

\noindent \textbf{Position to merge the layout embedding.} Table~\ref{t:layout-merge-layers} presents the impact of layout embedding fusion position on layout controllability. It can be seen that fusing layout embedding with the layers of denoising U-Net decoder can more effectively convey layout information.

\begin{table}
  \caption{Impact of pre-training dataset size on the quality of generated data, measured by AP$_{50}$. Pre-training diffusion models with more unlabeled images, especially those from the target domain, can improve data quality.}
  \label{t:pre-training-size}
  \centering
  \begin{tabular}{lcccccc}
    \toprule
    Dataset Size & w/o target images & w/ target images \\
    \midrule
    33k & 41.1 & 41.1 \\
    45k & 44.8 & 45.4 \\
    56k & 45.4 & 49.6 \\
    121k & 48.4 & 50.7\\
    \bottomrule
  \end{tabular}
\end{table}

\begin{table}
  \caption{Effectiveness of DODA-generated data on different detectors.}
  \label{t:detectors}
  \centering{
    \resizebox{\linewidth}{!}{
    \begin{tabular}{lccc}
    \toprule
    Method & Params & w/o DODA & w/ DODA \\
    \midrule
    Deformable DETR \cite{zhu2020deformable} & 41M & 10.4 & 37.2(\textcolor{ForestGreen}{+26.8})\\
    YOLOX L \cite{wang2023yolov7} & 54M & 30.5 & 50.7(\textcolor{ForestGreen}{+20.2}) \\
    YOLOV7 X \cite{wang2023yolov7} & 71M & 27.2 & 47.3(\textcolor{ForestGreen}{+20.1})\\
    FCOS X101 \cite{2019arXiv190401355T} & 90M & 20.1 & 41.5(\textcolor{ForestGreen}{+21.4}) \\
    \bottomrule
  \end{tabular}
  }}
\end{table}

\begin{table}[!t]
  \centering
  \begin{minipage}{0.5\linewidth}
  \caption{Ablations on domain encoder.}
  \makeatletter\def\@captype{table}
  \resizebox{\textwidth}{!}{
    \begin{tabular}{lc}
      \toprule
      Domain Encoder & FS$\uparrow$\\
      \midrule
      \xmark & 0.477 \\
      CLIP & \textbf{0.769} \\
      MAE & 0.747 \\
      ResNet101 & 0.751 \\
      \bottomrule
    \end{tabular}
    }
  \label{t:vision-encoders}
  \end{minipage}
  \hfill
  \begin{minipage}{0.43\linewidth}
    \caption{Features from different layers in MAE as domain embedding.}
    \makeatletter\def\@captype{table}
    \renewcommand\arraystretch{0.9}
    \resizebox{\textwidth}{!}{
      \begin{tabular}{lcc}
        \toprule
        Layers & FS$\uparrow$ & FID$\downarrow$ \\
        \midrule
        2 & 0.622 & 44.8 \\
        4 & 0.664 & 37.7 \\
        8 & 0.719 & 30.6 \\
        12 & 0.747 & 28.0 \\
        \bottomrule
      \end{tabular}
      }
    \label{t:mae-layers}
    \end{minipage}
\end{table}

\begin{table}
  \caption{Ablations on the position to merge the layout embedding.}
  \label{t:layout-merge-layers}
  \centering
  \resizebox{\linewidth}{!}{
    \begin{tabular}{ccccccccc}
      \toprule
      \multirow{2}{*}{Encoder} & \multirow{2}{*}{Decoder} & \multicolumn{6}{c}{YOLO Score\(\uparrow\)} & \multirow{2}{*}{FID\(\downarrow\)}  \\
      \cmidrule(lr){3-8}
      && mAP & AP\(_{50}\) & AP\(_{75}\) & AP\(^{s}\) & AP\(^{m}\) & AP\(^{l}\) \\
      \cmidrule(r){1-2}
      \cmidrule(r){3-9}
      \cmark & \xmark & 23.1 & 64.1 & 10.2 & 17.1 & 28.2 & 22.0 & 27.2\\
      \xmark & \cmark & 26.0 & 69.5 & 12.5 & 20.5 & 30.7 & 21.8 & 27.7\\
      \cmark & \cmark & 25.3 & 67.5 & 11.9 & 18.8 & 30.7 & 23.1 & 27.3\\
      \bottomrule
    \end{tabular}
  }
\end{table}
\section{Conclusion}
\label{sec:conclusion}

This paper presents DODA, a framework that incorporates domain features and image layout conditions to extend a diffusion model, enabling it to generate detection data for new agricultural domains. With just a few reference images from the target domain, DODA can generate data for it without training. Extensive experiments demonstrated the effectiveness of DODA-generated data in adapting detectors to diverse agricultural domains, as shown by significant AP improvements across multiple domains. The simplicity and effectiveness of DODA reduce barriers for more growers to use object detection for their personalized scenarios.
\section*{Acknowledgments}

This work was supported by JSPS KAKENHI Grant Number 25H01110 and the Hokkaido Sarabetsu Village “Endowed Chair for Field Phenomics” projects in Japan.

{
    \small
    \bibliographystyle{ieeenat_fullname}
    \bibliography{ref}
}
\appendix
\onecolumn
\section*{Appendix}

\section{Derivation of Training Objective}
\label{app:derivation}
Based on the assumption that the diffusion model can learn to utilize conditions during training, thereby generating images $\hat{\mathbf{x}} \sim p(\mathbf{x} | \mathbf{y}_1, \mathbf{y}_2, \dots, \mathbf{y}_n)$, many studies integrate multiple conditions, e.g., text, depth, pose, into the training of diffusion model \cite{li2023open, chen2023integrating, lu2023handrefiner}:
\begin{equation}
  \label{eqn:opt_cgm_2condition}
    \begin{aligned}
    {\boldsymbol{\theta}}^* = \argmin_{\boldsymbol{\theta}} \,
    & \mathbb{E}_{t \sim U(0, T)} \,
    \mathbb{E}_{\mathbf{x}_0,\mathbf{y}_1,\mathbf{y}_2 \sim p(\mathbf{x}_0,\mathbf{y}_1,\mathbf{y}_2)} \,
    \mathbb{E}_{\mathbf{x}_t \sim p(\mathbf{x}_t|\mathbf{x}_0)}
    [\lambda(t)\| s(\mathbf{x}_t,\mathbf{y}_1,\mathbf{y}_2,t;{\boldsymbol{\theta}}) 
    - \nabla_{\mathbf{x}_t}\log p(\mathbf{x}_t|\mathbf{x}_0) \|^2 ]
    \end{aligned}
\end{equation}
However, these studies simply incorporate multiple conditions into the diffusion model without modifying the optimization objective, making it unclear whether $\hat{\mathbf{x}} \sim p(\mathbf{x} | \mathbf{y}_1, \mathbf{y}_2, \dots, \mathbf{y}_n)$ is actually achieved. To address this, we derive the two-conditional optimization objective as follows.

First, applying the forward diffusion process in Eq.~\ref{eqn:forward_sde}, obtains the perturbed distribution $p(\mathbf{x}_t | \mathbf{y}_1, \mathbf{y}_2)$, according to \cite{anderson1982reverse}, the corresponding reverse-time SDE is given by:
\begin{equation}
  \label{eqn:backward_sde_2condition}
  d\mathbf{x} = [f(\mathbf{x}, t) - g(t)^2  \nabla_{\mathbf{x}_t} \log p(\mathbf{x}_t | \mathbf{y}_1, \mathbf{y}_2)] dt + g(t) d \bar{\mathbf{w}}
\end{equation}
By simulating Eq.~\ref{eqn:backward_sde_2condition}, we can generate samples from $p(\mathbf{x}_t | \mathbf{y}_1, \mathbf{y}_2)$. To construct the reverse-time SDE, we need to estimate the conditional score. Similar to Eq.~\ref{eqn:opt_sgm}, the training objective is:
\begin{equation}
  \label{eqn:opt_sgm_2condition}
  \begin{aligned}
  {\boldsymbol{\theta}}^* = \argmin_{\boldsymbol{\theta}} \,
  & \mathbb{E}_{t \sim U(0, T)} \,
  \mathbb{E}_{\mathbf{x}_t,\mathbf{y}_1,\mathbf{y}_2 \sim p(\mathbf{x}_t,\mathbf{y}_1,\mathbf{y}_2)}
  [\lambda(t)\| s(\mathbf{x}_t,\mathbf{y}_1,\mathbf{y}_2,t;{\boldsymbol{\theta}}) 
  - \nabla_{\mathbf{x}_t}\log p(\mathbf{x}_t|\mathbf{y}_1,\mathbf{y}_2) \|^2 ]
  \end{aligned}
\end{equation}

However, the $\nabla_{\mathbf{x}_t}\log p(\mathbf{x}_t|\mathbf{y}_1,\mathbf{y}_2)$ in Eq.~\ref{eqn:opt_sgm_2condition} is hard to access. \cite{batzolis2021conditional} provided a method to approximate $\nabla_{\mathbf{x}_t}\log p(\mathbf{x}_t|\mathbf{y})$. By generalizing it to multi-conditional setting, we prove that the optimal solution of Eq.~\ref{eqn:opt_sgm_2condition} is the same as the solution of Eq.~\ref{eqn:opt_cgm_2condition}:

\begin{Proposition_proof}
  \label{prop:2condition}
  The solution that minimizes \\
  $\mathbb{E}_{t \sim U(0, T)} 
  \mathbb{E}_{\mathbf{x}_0,\mathbf{y}_1,\mathbf{y}_2 \sim p(\mathbf{x}_0,\mathbf{y}_1,\mathbf{y}_2)} 
  \mathbb{E}_{\mathbf{x}_t \sim p(\mathbf{x}_t|\mathbf{x}_0)} 
  [\lambda(t)\| s(\mathbf{x}_t,\mathbf{y}_1,\mathbf{y}_2,t;{\boldsymbol{\theta}}) - \nabla_{\mathbf{x}_t}\log p(\mathbf{x}_t|\mathbf{x}_0) \|^2 ]$ \\
  is the same as the solution minimizes \\
  $\mathbb{E}_{t \sim U(0, T)} 
  \mathbb{E}_{\mathbf{x}_t,\mathbf{y}_1,\mathbf{y}_2 \sim p(\mathbf{x}_t,\mathbf{y}_1,\mathbf{y}_2)} 
  [\lambda(t)\| s(\mathbf{x}_t,\mathbf{y}_1,\mathbf{y}_2,t;{\boldsymbol{\theta}}) - \nabla_{\mathbf{x}_t}\log p(\mathbf{x}_t|\mathbf{y}_1,\mathbf{y}_2) \|^2 ]$
\end{Proposition_proof}

\begin{proof}
  Let $f(\mathbf{x}_t,\mathbf{x}_0,\mathbf{y}_1,\mathbf{y}_2) := \lambda(t)\| s(\mathbf{x}_t,\mathbf{y}_1,\mathbf{y}_2,t;{\boldsymbol{\theta}}) - \nabla_{\mathbf{x}_t}\log p(\mathbf{x}_t|\mathbf{x}_0) \|^2 $, first, according to the Law of Iterated Expectations, we have:
  \begin{align}
    \label{eqn:proof1}
    &   \mathbb{E}_{t \sim U(0, T)} 
    \mathbb{E}_{\mathbf{x}_0,\mathbf{y}_1,\mathbf{y}_2 \sim p(\mathbf{x}_0,\mathbf{y}_1,\mathbf{y}_2)} 
    \mathbb{E}_{\mathbf{x}_t \sim p(\mathbf{x}_t|\mathbf{x}_0)} 
    [\lambda(t)\| s(\mathbf{x}_t,\mathbf{y}_1,\mathbf{y}_2,t;{\boldsymbol{\theta}}) - \nabla_{\mathbf{x}_t}\log p(\mathbf{x}_t|\mathbf{x}_0) \|^2 ] \notag \\
    =&  \mathbb{E}_{t \sim U(0, T)} 
    \mathbb{E}_{\mathbf{y}_1,\mathbf{y}_2 \sim p(\mathbf{y}_1,\mathbf{y}_2)} 
    \mathbb{E}_{\mathbf{x}_0 \sim p(\mathbf{x}_0|\mathbf{y}_1,\mathbf{y}_2)} 
    \mathbb{E}_{\mathbf{x}_t \sim p(\mathbf{x}_t|\mathbf{x}_0)} 
    [f(\mathbf{x}_t,\mathbf{x}_0,\mathbf{y}_1,\mathbf{y}_2)] \notag \\
    =&  \mathbb{E}_{t \sim U(0, T)} 
    \mathbb{E}_{\mathbf{y}_2 \sim p(\mathbf{y}_2)} 
    \mathbb{E}_{\mathbf{y}_1 \sim p(\mathbf{y}_1|\mathbf{y}_2)} 
    \mathbb{E}_{\mathbf{x}_0 \sim p(\mathbf{x}_0|\mathbf{y}_1,\mathbf{y}_2)} 
    \mathbb{E}_{\mathbf{x}_t \sim p(\mathbf{x}_t|\mathbf{x}_0)} 
    [f(\mathbf{x}_t,\mathbf{x}_0,\mathbf{y}_1,\mathbf{y}_2)]
  \end{align}
  The $\mathbf{y}_1$ and $\mathbf{y}_2$ are independent of each other. Given $\mathbf{x}_0$, $\mathbf{y}_1$ and $\mathbf{y}_2$ are independent of $\mathbf{x}_t$. Let $g(\mathbf{x}_t,\mathbf{x}_0,\mathbf{y}_1,\mathbf{y}_2) := \lambda(t)\| s(\mathbf{x}_t,\mathbf{y}_1,\mathbf{y}_2,t;{\boldsymbol{\theta}}) - \nabla_{\mathbf{x}_t}\log p(\mathbf{x}_t|\mathbf{x}_0,\mathbf{y}_1,\mathbf{y}_2) \|^2  $, Eq.~\ref{eqn:proof1} can be written as:
  \begin{align}
    \label{eqn:proof2}
    &  \mathbb{E}_{t \sim U(0, T)} 
    \mathbb{E}_{\mathbf{y}_2 \sim p(\mathbf{y}_2)} 
    \mathbb{E}_{\mathbf{y}_1 \sim p(\mathbf{y}_1)} 
    \mathbb{E}_{\mathbf{x}_0 \sim p(\mathbf{x}_0|\mathbf{y}_1,\mathbf{y}_2)} 
    \mathbb{E}_{\mathbf{x}_t \sim p(\mathbf{x}_t|\mathbf{x}_0)} 
    [f(\mathbf{x}_t,\mathbf{x}_0,\mathbf{y}_1,\mathbf{y}_2)] \notag \\
    =& \mathbb{E}_{t \sim U(0, T)} 
    \mathbb{E}_{\mathbf{y}_2 \sim p(\mathbf{y}_2)} 
    \mathbb{E}_{\mathbf{y}_1 \sim p(\mathbf{y}_1)} 
    \mathbb{E}_{\mathbf{x}_0 \sim p(\mathbf{x}_0|\mathbf{y}_1,\mathbf{y}_2)} 
    \mathbb{E}_{\mathbf{x}_t \sim p(\mathbf{x}_t|\mathbf{x}_0,\mathbf{y}_1,\mathbf{y}_2)} 
    [g(\mathbf{x}_t,\mathbf{x}_0,\mathbf{y}_1,\mathbf{y}_2)]
  \end{align}
  Let $t$, $\mathbf{y}_1$ and $\mathbf{y}_2$ be arbitrary fixed values, then we can define $h(\mathbf{x}_t) := s(\mathbf{x}_t,\mathbf{y}_1,\mathbf{y}_2,t;{\boldsymbol{\theta}})$, $q(\mathbf{x}_0) := p(\mathbf{x}_0|\mathbf{y}_1,\mathbf{y}_2)$ and $q(\mathbf{x}_t|\mathbf{x}_0) := p(\mathbf{x}_t|\mathbf{x}_0,\mathbf{y}_1,\mathbf{y}_2)$, applying the Law of Iterated Expectations, we have:
  \begin{align}
    \label{eqn:proof3}
    &   \mathbb{E}_{\mathbf{x}_0 \sim p(\mathbf{x}_0|\mathbf{y}_1,\mathbf{y}_2)} 
    \mathbb{E}_{\mathbf{x}_t \sim p(\mathbf{x}_t|\mathbf{x}_0,\mathbf{y}_1,\mathbf{y}_2)} 
    [\lambda(t)\| s(\mathbf{x}_t,\mathbf{y}_1,\mathbf{y}_2,t;{\boldsymbol{\theta}}) - \nabla_{\mathbf{x}_t}\log p(\mathbf{x}_t|\mathbf{x}_0,\mathbf{y}_1,\mathbf{y}_2) \|^2 ] \notag \\
    =&  \mathbb{E}_{\mathbf{x}_0 \sim q(\mathbf{x}_0)} 
    \mathbb{E}_{\mathbf{x}_t \sim q(\mathbf{x}_t|\mathbf{x}_0)} 
    [\lambda(t)\| h(\mathbf{x}_t) - \nabla_{\mathbf{x}_t}\log q(\mathbf{x}_t|\mathbf{x}_0) \|^2 ] \notag \\
    =&  \mathbb{E}_{\mathbf{x}_t \sim q(\mathbf{x}_t)} 
    [\lambda(t)\| h(\mathbf{x}_t) - \nabla_{\mathbf{x}_t}\log q(\mathbf{x}_t) \|^2 ] \notag \\
    =&  \mathbb{E}_{\mathbf{x}_t \sim p(\mathbf{x}_t|\mathbf{y}_1,\mathbf{y}_2)} 
    [\lambda(t)\| s(\mathbf{x}_t,\mathbf{y}_1,\mathbf{y}_2,t;{\boldsymbol{\theta}}) - \nabla_{\mathbf{x}_t}\log p(\mathbf{x}_t|\mathbf{y}_1,\mathbf{y}_2) \|^2 ]
  \end{align}
  Since $t$, $\mathbf{y}_1$, $\mathbf{y}_2$ are arbitrary, Eq.~\ref{eqn:proof3} is true for all $t$, $\mathbf{y}_1$, $\mathbf{y}_2$, via Eq.~\ref{eqn:proof3} and the Law of Iterated Expectations, we can easily rewrite Eq.~\ref{eqn:proof2} as:
  \begin{align}
    \label{eqn:proof4}
    &   \mathbb{E}_{t \sim U(0, T)} 
    \mathbb{E}_{\mathbf{y}_2 \sim p(\mathbf{y}_2)} 
    \mathbb{E}_{\mathbf{y}_1 \sim p(\mathbf{y}_1)} 
    \mathbb{E}_{\mathbf{x}_t \sim p(\mathbf{x}_t|\mathbf{y}_1,\mathbf{y}_2)} 
    [\lambda(t)\| s(\mathbf{x}_t,\mathbf{y}_1,\mathbf{y}_2,t;{\boldsymbol{\theta}}) - \nabla_{\mathbf{x}_t}\log p(\mathbf{x}_t|\mathbf{y}_1,\mathbf{y}_2) \|^2 ] \notag \\
    =&  \mathbb{E}_{t \sim U(0, T)} 
    \mathbb{E}_{\mathbf{x}_t,\mathbf{y}_1,\mathbf{y}_2 \sim p(\mathbf{x}_t,\mathbf{y}_1,\mathbf{y}_2)} 
    [\lambda(t)\| s(\mathbf{x}_t,\mathbf{y}_1,\mathbf{y}_2,t;{\boldsymbol{\theta}}) - \nabla_{\mathbf{x}_t}\log p(\mathbf{x}_t|\mathbf{y}_1,\mathbf{y}_2) \|^2 ]
  \end{align}
\end{proof}

\noindent With this Proposition, we have established that the optimal solution $s(\mathbf{x}_t,\mathbf{y}_1,\mathbf{y}_2,t;{{\boldsymbol{\theta}}^*})$ of Eq.~\ref{eqn:opt_cgm_2condition} is able to approximate the multi-conditional score $\nabla_{\mathbf{x}_t}\log p(\mathbf{x}_t|\mathbf{y}_1,\mathbf{y}_2)$.

\section{Algorithm}
\label{alg:greedycoloring}
\begin{algorithm}[H]    
  \caption{Bounding Boxes Arrangement}
  
  \textbf{Input}: Adjacency matrix $A$ of bounding boxes, the number of bounding boxes $n$.\\
  \textbf{Output}: Array $channels$ containing the assigned channel for each bounding box.

  \begin{algorithmic}[1]
    \STATE $channels \leftarrow \text{array of length } n \text{ initialized with } 0$
    \FOR{ $i = 1$ to $n$}
        \STATE $C_i \leftarrow \emptyset$
        \FOR{$j = 1$ to $i$}
            \IF{$A[i][j] = 1$ and $channels[j] \neq 0$}
                \STATE Add $channels[j]$ to $C_i$
            \ENDIF
        \ENDFOR
        \STATE Assign the smallest channel not in $C_i$ to $channels[i]$
    \ENDFOR
  \end{algorithmic}
  \textbf{Return} $channels$
\end{algorithm}

\section{Implementation details}
\label{app:imple}
\textbf{Dataset preparation.} In the GWHD, the images have high resolution but are relatively few in number. Therefore, we divided the original \( 1024 \times 1024 \) images into 9 images of size \( 512 \times 512 \) with step size 256. After splitting, there are a total of 58,635 images in GWHD. For the COCO 2017 dataset, we train with the official training set and test the proposed L2I method on the validation set.

\noindent \textbf{Domain adaption.} To evaluate the effectiveness of DODA for domain adaptation, we focus on the domains within the GWHD test set where AP$_{50}$ lower than 0.8. For each domain, we use DODA to generate a 200 image dataset, then fine-tune the YOLOX-L on this synthetic data for one epoch.  

\noindent \textbf{L2I on COCO.} Following the setting of \cite{chen2023integrating, cheng2023layoutdiffuse}, we apply the proposed LI2I method to Stable Diffusion \cite{rombach2022high} v1.5. To preserve the knowledge learned from billions of images \cite{schuhmann2022laion}, we use the encoder of U-Net as the layout encoder, and following \cite{zhang2023adding} initialize it with the weight of the diffusion model. Since Stable Diffusion is a T2I model, we constructed a simple text prompt for our method: "a photograph with $(N_{\text{cls}}^1) ({Cls}^1), \ldots, (N_{{cls}}^i) ({Cls}^i)$", where $Cls^i$ is the category, and $N_{cls}^i$ denotes the number of objects belonging to that category. Since COCO contains multiple categories, we design a layout coding method that different from Sec.~\ref{sec:layout_encoder}, objects of the same category are depicted with the same hue but weaker brightness, and the bounding box of each object is drawn in descending order of area.

\noindent \textbf{Hyperparameters.} By default, we use 4 NVIDIAA-V100-32GB, but all models in this paper can be trained on one single V100, and the GPU Memory usage and approximate computational requirements for one GPU are provided in the last two rows of Table~\ref{Hyper1} and Table~\ref{Hyper2}. When training with multiple cards, all parameters including Learning Rate are the same, except Iterations.

\begin{table}[htbp]
  \caption{Hyperparameters for pre-training DODA. DODA leverages latent diffusion (LDM) \cite{rombach2022high} as the base diffusion model, which uses variational autoencoder (VAE) \cite{kingma2013auto} to encode the image into the latent space and thus reduces the computation, so the pre-training of DODA is divided into two stages: the VAE and LDM.}
  \label{Hyper1}
  \centering
    \begin{tabular}{lccc}
    \toprule
      \multicolumn{2}{l}{ } & VAE & LDM \\
    \midrule
      \multicolumn{2}{l}{Dataset} & All images in GWHD & All images in GWHD \\
      \multicolumn{2}{l}{Target Image Shape} & \(256 \times 256 \times 3\) & \(256 \times 256 \times 3\) \\
      \multicolumn{2}{l}{Domain Reference Image Shape} & - & \(224 \times 224 \times 3\) \\
    \midrule
      \multirow{4}{*}{Data Augmentation} & \multirow{3}{*}{Target Image} & Random Rotation & Random Rotation \\
        && Random Crop & Random Crop\\
        && Random Flip & Random Flip\\
      \cmidrule(lr){2-4}
      & Reference Image & - & Random Crop \\
    \midrule
      \multicolumn{2}{l}{f} & 4 & 4 \\
      \multicolumn{2}{l}{Channels} & 128 & 224 \\
      \multicolumn{2}{l}{Channel Multiplier} & 1,2,4 & 1,2,4 \\
      \multicolumn{2}{l}{Attention Resolutions} & - & 2,4 \\
      \multicolumn{2}{l}{Number of Heads} & - & 8 \\
    \midrule
      \multicolumn{2}{l}{Learning Rate} & 2.5e-6 & 2.5e-5\\
      \multicolumn{2}{l}{Iterations} & 480k & 600k \\
      \multicolumn{2}{l}{Batch Size} & 8 & 16 \\
    \midrule
      \multicolumn{2}{l}{GPU Memory usage} & 32 GB & 16 GB \\
      \multicolumn{2}{l}{Computational consumption} & 20 v100-days & 14 v100-days \\
    \bottomrule

  \end{tabular}
\end{table}

\begin{table}[htbp]
  \caption{Hyperparameters for layout-to-image.}
  \label{Hyper2}
  \centering
  {
    \begin{tabular}{@{}lcccc@{}}
    \toprule
      \multicolumn{2}{l}{Dataset} & COCO 2017 training & COCO 2017 training & GWHD training \\
    \midrule
      \multicolumn{2}{l}{Target/Layout Image Shape} & \(256 \times 256 \times 3\) & \(512 \times 512 \times 3\) & \(256 \times 256 \times 3\) \\
      \multicolumn{2}{l}{Domain Reference Image Shape} & - & - & \(224 \times 224 \times 3\) \\
    \midrule
      \multirow{4}{*}{Data Augmentation} & \multirow{3}{*}{Target Image} & Random Flip & Random Flip & Random Rotation \\
        &&&& Random Crop \\
        &&&& Random Flip\\
      \cmidrule(lr){2-5}
      & Reference Image & - & - & Random Crop \\
    \midrule
      \multicolumn{2}{l}{Base Model} & SD1.5 & COCO 256 & LDM in Table~\ref{Hyper1} \\
      \multicolumn{2}{l}{f} & 8 & 8 & 4 \\
      \multicolumn{2}{l}{Channels} & 320 & 320 & 224 \\
      \multicolumn{2}{l}{Channel Multiplier} & 1,2,4,4 & 1,2,4,4 & 1,2,4 \\
      \multicolumn{2}{l}{Attention Resolutions} & 1,2,4 & 1,2,4 & 2,4 \\
      \multicolumn{2}{l}{Number of Heads} & 8 & 8 & 8 \\
    \midrule
      \multicolumn{2}{l}{Learning Rate} & 2.5e-5 & 2.5e-5 & 1e-5 \\
      \multicolumn{2}{l}{Iterations} & 100K & 30K & 80K \\
      \multicolumn{2}{l}{Batch Size} & 16 & 8 & 16 \\
    \midrule
    \multicolumn{2}{l}{GPU Memory usage} & 27 GB & 25 GB & 20 GB \\
    \multicolumn{2}{l}{Computational consumption} & 40 v100-hours & 56 v100-hours & 40 v100-hours \\
    \bottomrule
  \end{tabular}
  }
\end{table}

\section{Evaluation metrics}
\label{app:eval_metric}
\textbf{Fréchet Inception Distance (FID)} \cite{heusel2017gans} reflects the quality of the generated image. FID measures similarity of features between two image sets and the features extracted by the pre-trained Inception-V3 \cite{szegedy2016rethinking}.

\noindent \textbf{Inception Score (IS)} \cite{salimans2016improved} uses a pre-trained Inception-V3 \cite{szegedy2016rethinking} to classify the generated images, reflecting the diversity and quality of the images. When calculating the IS for Table~\ref{t:coco-L2I}, as in the original paper, we divided the data into 10 splits. The error bar for IS is the standard deviation between the splits.

\noindent \textbf{COCO Metrics} refers to fine-tuning detectors using synthetic data, and then calculating AP according to the official COCO.

\noindent \textbf{YOLO Score} uses a pre-trained YOLOX-L \cite{ge2021yolox} to detect the generated image, and calculates the AP between the detection result and the input label, which reflects the ability of the generated model to control the layout.

\noindent \textbf{Feature Similarity (FS)}. As discussed in Sec.\ref{M:DA}, domain shift manifests in feature differences, the domain encoder should guide diffusion to generate images aligned with reference images' features. Here we use DINO-V2 \cite{oquab2023dinov2} to extract features from the generated images and their corresponding reference images, calculate the cosine similarity for each pair, and then compute the average similarity across multiple image pairs. Compared with FID, FS provides more fine-grained information.

\section{More Ablation}
\label{app:more_AB}

\noindent \textbf{Channel coding.} Table~\ref{t:channel-coding} evaluates the effectiveness of our channel coding component. By representing overlapping instances through different color channels, the model better distinguish overlapped instances and thus more accurately control the layout.

\noindent \textbf{Reference image selection method.} In the main experiments, we randomly sample reference images $x_{ref}$ from the entire set of target domain images $x$. In this section, we investigate how the choice of reference images affects the generated data. We first sample different numbers of images from $x$ to create reference pools $x_{pool}^i$ of varying sizes, and then randomly sample $x_{ref}$ from each $x_{pool}^i$. For each $x_{pool}^i$, we repeat the sampling process 5 times to compute the standard deviation. As shown in Table~\ref{t:choose_ref}, when the reference pool is extremely small, the diversity of $x_{ref}$ is low, resulting in low AP scores, and the standard deviation is large because the sampling bias is amplified. Once the size of the $x_{pool}^i$ exceeds 100, the AP stabilizes.

\begin{figure}
  \centering
  \includegraphics[width=\linewidth]{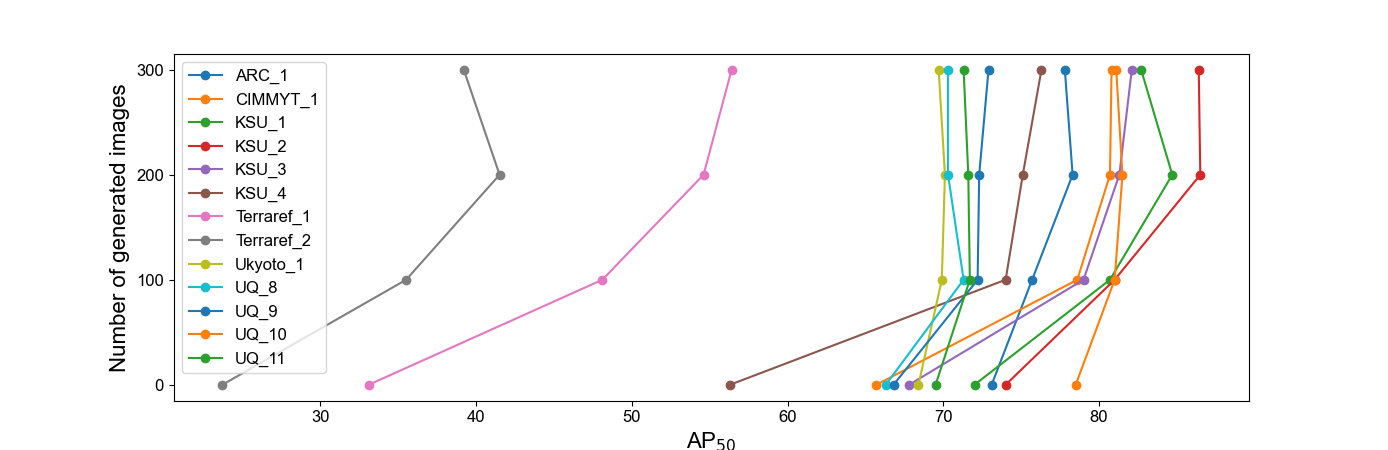}
  \caption{Ablations on the number of generated images. For most domains, 200 generated images are sufficient.}
  \label{fig:n_img}
\end{figure}

\noindent \textbf{Number of generated images.} We investigate changes in the performance of using different amounts of generated data. As shown in Fig.~\ref{fig:n_img}, for most domains, a dataset consisting of 200 synthetic images is sufficient to convey the characteristics of the target domain. Increasing the number of images does not significantly improve performance. To ensure consistency across experiments, we use 200 images by default for all domains.

\begin{table}[H]
  \caption{Ablations on the layout channel coding. Channel coding can help the model more accurately control layout.}
  \label{t:channel-coding}
  \centering{
    \setlength{\extrarowheight}{-1pt}
    \begin{tabular}{ccccccc}
      \toprule
      \multirow{2}{*}{Channel coding} & \multicolumn{6}{c}{YOLO Score\(\uparrow\)} \\
      \cmidrule(lr){2-7}
      & mAP & AP\(_{50}\) & AP\(_{75}\) & AP\(^{s}\) & AP\(^{m}\) & AP\(^{l}\) \\
      \midrule
      \xmark & 26.4 & 67.8 & 14.5 & 20.0 & 31.3 & 28.6\\
      \cmark & 27.4 & 70.0 & 15.3 & 20.8 & 32.7 & 29.9\\
      \bottomrule
    \end{tabular}
    }
\end{table}

\begin{table}[H]
  \caption{Ablations on the selecting of reference images.}
  \label{t:choose_ref}
  \centering{
    \setlength{\extrarowheight}{-1pt}
    \begin{tabular}{ccc}
    \toprule
    References pool size & AP\(_{50}\)\\
    \midrule
    0 & 30.5 \\
    10 & 37.22\(\pm\)8.89 \\
    100 & 48.00\(\pm\)1.92 \\
    400 & 49.36\(\pm\)1.55 \\
    1600 & 48.58\(\pm\)1.95 \\
    \bottomrule
  \end{tabular}
  }
\end{table}

\section{Influence of Noisy Samples in Pre-training Data}

\begin{figure}[htbp]
  \centering
  \includegraphics[width=\linewidth]{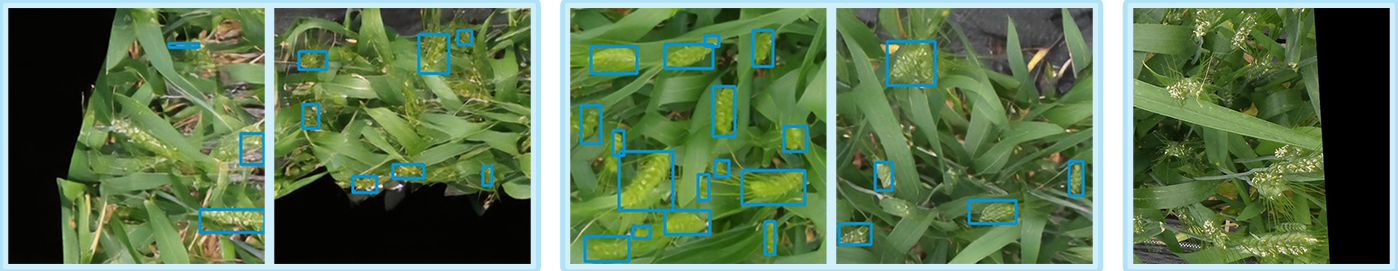}
  \caption{Examples of generated images in domain ``Ukyoto\_1''. Left, many generated images have unnatural black edges. Middle, normal generated images, which are better aligned with the input layout (blue bounding boxes) than the images with black edges. Right, some real images used for pre-training also have black edges.}
  \label{fig:Ukyoto}
\end{figure}

On the ``Ukyoto\_1'' domain, the improvement in mAP is only marginal. As shown in Fig.~\ref{fig:Ukyoto} left, we observed many images with an unusual black area. Compared to images without black edges (Fig.~\ref{fig:Ukyoto} middle), these images also show poorer alignment with the given layout. The black regions and the poorer alignment degrade the quality of generated data. Upon further inspection, we found that these black regions originate from the real images used for pre-training (as illustrated in Fig.~\ref{fig:Ukyoto} right). When preparing the pre-training dataset, it necessary to filter out such images to improve data quality.

\section{Qualitative Results of Agricultural Object Detection Domain Adaptation}
\label{app:qualitative_da}
\begin{figure}[htbp]
  \centering
  \includegraphics[width=0.8\linewidth]{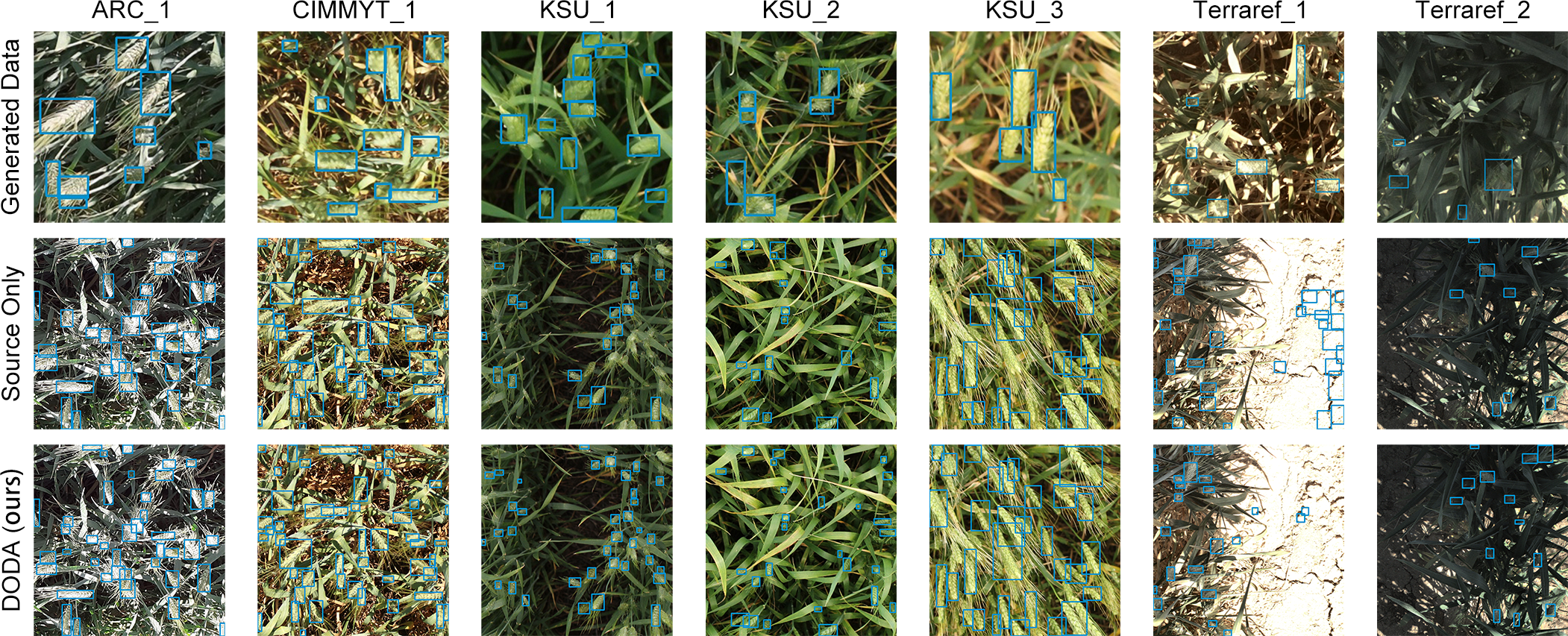}
  \caption{Visualization of generated data and detection results on target domains.}
  \label{fig:DA_visualization}
\end{figure}

\clearpage

\section{Qualitative comparisons with previous L2I methods on COCO}
\label{app:qualitative_comparisons_coco}
\begin{figure}[htbp]
  \centering
  \includegraphics[width=0.85\linewidth]{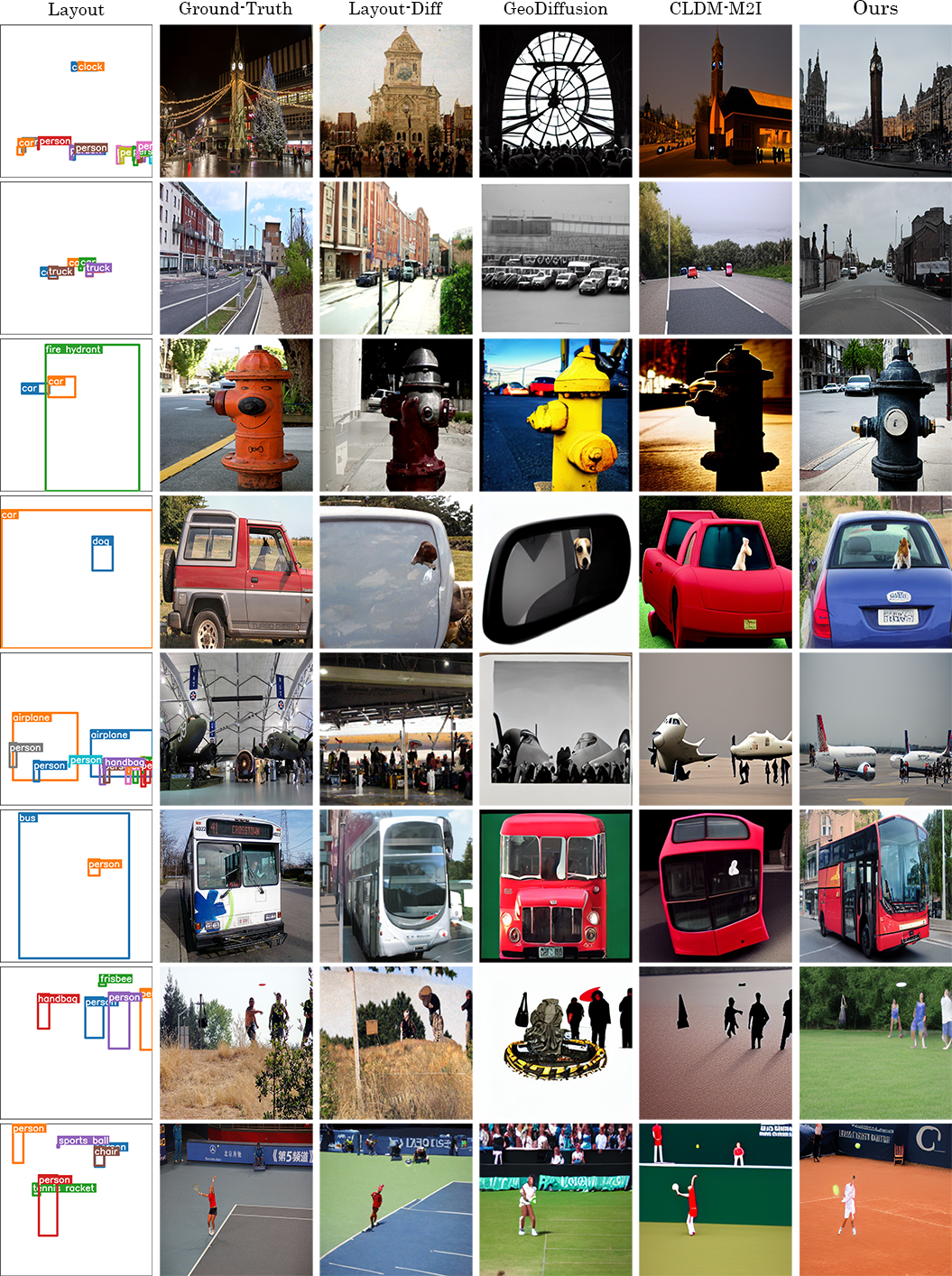}
  \caption{Visualization of comparisons between our proposed LI2I method and previous LT2I methods on COCO. LI2I generates images with more detail and greater control over layout, especially for small objects.}
  \label{fig:coco-example}
\end{figure}

\end{document}